\documentclass[runningheads]{llncs}
\usepackage[T1]{fontenc}
\usepackage{graphicx, amsmath, multirow, siunitx, hyperref}
\usepackage{color}

\begin{document}
\title{Beta-Sigma VAE: Separating beta and decoder variance in Gaussian variational autoencoder}
\titlerunning{Beta-Sigma VAE}
%
\author{Seunghwan Kim\orcidID{0009-0004-3373-5991} \and
Seungkyu Lee\orcidID{0000-0002-9721-4093}}
\authorrunning{S. Kim and S. Lee}
%
\institute{Dept. of Computer Science and Engineering, Kyung Hee University, South Korea
\email{\{overnap,seungkyu\}@khu.ac.kr}}
\maketitle              
\begin{abstract}
Variational autoencoder (VAE) is an established generative model but is notorious for its blurriness. In this work, we investigate the blurry output problem of VAE and resolve it, exploiting the variance of Gaussian decoder and $\beta$ of beta-VAE~\cite{higgins2016beta}. Specifically, we reveal that the indistinguishability of decoder variance and $\beta$ hinders appropriate analysis of the model by random likelihood value, and limits performance improvement by omitting the gain from $\beta$. To address the problem, we propose Beta-Sigma VAE (BS-VAE) that explicitly separates $\beta$ and decoder variance $\sigma^2_x$ in the model. Our method demonstrates not only superior performance in natural image synthesis but also controllable parameters and predictable analysis compared to conventional VAE. In our experimental evaluation, we employ the analysis of rate-distortion curve and proxy metrics on computer vision datasets. The code is available on \url{https://github.com/overnap/BS-VAE}.

\end{abstract}

\section{Introduction}

Generative modeling has been a headliner of deep learning research over the last decade. It approximates the distribution of observed samples such as natural images or natural language sentences. Variational autoencoder (VAE)~\cite{kingma2013auto,rezende2014stochastic}, one of the most popular generative deep neural networks with well-developed mathematical background, has demonstrated competitive performance in realistic sample synthesis~\cite{razavi2019vq2,bowman2016text}, image segmentation~\cite{kohl2018segmentation}, data augmentation~\cite{norouzi2020augmentation}, image compression~\cite{duan2023compress2}, and reinforcement learning~\cite{nair2018reinforcement,pong2020skew}.

However, VAE has a notorious blurry output problem that hinders achieving cutting-edge generation quality. As a consequence, VAE has been adopted in various downstream tasks, but left off in major generative network applications. The technical source of the blurry output problem is difficult to pinpoint. Prior methods have been proposed to improve either the reconstruction quality or generation quality of VAEs with the variance of decoder distribution~\cite{rybkin2021simple} and $\beta$ of beta-VAE~\cite{higgins2016beta}. The lower the variance of decoder is, the sharper the output images are, since the variance represents the noise of decoder distribution. In return, the risk of bad local minimizers increases, as the loss smoothing effect of high variance is reduced~\cite{dai2021value}. On the other hand, $\beta$ extends VAE outside the likelihood, which allows beta-VAE to obtain useful properties such as latent disentanglement~\cite{higgins2016beta,burgess2018understanding,chen2018isolating,esmaeili2019structured} and rate-distortion tradeoff~\cite{alemi2018fixing,bozkurt2021rate,bae2022multi}. One can achieve sharp output by carefully tuning $\beta$.

These two parameters appear to have similar effects. Moreover, in special cases, e.g. Gaussian VAE with constant decoder variance, they are mathematically equivalent. Nevertheless, as they have separate design motivations, it is clear that their purposes and impacts are different. Confusion with the two parameters in prior approaches hinder performance improvement and model analysis of VAEs. For example, a method considering the two parameters are the same and optimizing a single integrated parameter cannot achieve the optimality of two parameters properly. The integrated parameter also leads to an indeterminate variance, so the likelihood value becomes arbitrary. In this case, likelihood values can vary for the same model and weights making the comparison virtually meaningless, which is very damaging to the research of VAEs.

In this work, we analyze the confusion about the influence of decoder variance and $\beta$, and propose a simple solution that derives optimal performance of VAEs.

Our contributions are as follows:
\begin{itemize}
    \item \textbf{Investigation of blurry output problem in VAEs.} The blurry output is a complex problem that is difficult to explain with any single factor. We classify it into poor reconstruction and poor generation followed by respective problem definitions and analysis.
    
    \item \textbf{Identification of the problems occurring in Gaussian VAE in which the variance of decoder $\sigma^2_x$ and $\beta$ of beta-VAE~\cite{higgins2016beta} are considered as a single integrated parameter.} Both parameters show similar effects and have been used to address the blurry output problem of VAEs. On the other hand, based on their different design motivations, $\sigma^2_x$ and $\beta$ affect the quality of reconstruction and generation respectively, which introduces non-optimality in the performance of VAEs.
      
    \item \textbf{Proposing a simple and explicit method to separate $\beta$ and $\sigma^2_x$.} Our method, Beta-Sigma VAE (BS-VAE), improves the performance of Gaussian VAE, as it takes advantage of both parameters. It also makes VAE more controllable, since it obtains a model of the rate-distortion curve with optimal decoder variance. Furthermore, it ensures that the same model and weights always have the same likelihood value, which enables predictable and meaningful analysis.
\end{itemize}

Our claims are validated on computer vision datasets. Our method, BS-VAE, is independent of architecture and scale, so it is applicable to most VAE-variants. We hope that our efforts encourage following research on VAEs to extend constructive analysis and accomplish competitive performance in many generative network applications.

\section{Background}

\textbf{Variational autoencoder (VAE).} VAE~\cite{kingma2013auto,rezende2014stochastic} models a parameterized distribution $p_\theta(x) = \int p_\theta(x|z)p(z)dz$ for the observable variable $x$ and latent variable $z$. It is fundamentally a maximum likelihood estimation. The log-likelihood $\log p_\theta(x)$ is generally intractable. Hence, VAE performs variational inference employing variational distribution $q_\phi(z|x)$. It learns evidence lower bound (ELBO) of the log-likelihood that consists of reconstruction error, Equation~\eqref{elbo:recon}, and KL divergence, Equation~\eqref{elbo:kl}. Note that the objectives are about a single sample $x_i$ for convenience.

\begin{align}
    &-\log{p_\theta(x_i)} \leq -\text{ELBO}(\theta, \phi, x_i) \nonumber \\
    = &-E_{z\sim q_\phi(z|x_i)}[\log{p_\theta(x_i|z)}] \label{elbo:recon} \\
    &+ D_{KL}(q_\phi (z|x_i)||p(z)) \label{elbo:kl}
\end{align}

\textbf{Gaussian VAE.} The architecture of VAE, the encoder $q_\phi(z|x)$ followed by the decoder $p_\theta(x|z)$, is similar to an autoencoder. Different from autoencoder, VAE establishes probability distributions which are usually set to Gaussian in computer vision applications~\cite{kingma2013auto,sonderby2016ladder,dai2018diagnosing}. For the observable variable $x$ and latent variable $z$, Gaussian VAE is the variational autoencoder consisting of the following encoder $q_\phi(z|x)$ and decoder $p_\theta(x|z)$.
\begin{gather*}
    q_\phi(z|x) \sim \mathcal{N}(\mu_z(x),\Sigma_z(x)) \\
    p_\theta(x|z) \sim \mathcal{N}(\mu_x(z), \Sigma_x(z))
\end{gather*}
where $\Sigma_z$ is the diagonal covariance matrix and $\Sigma_x$ is the scalar matrix in conventional setting.
\begin{align*}
    \Sigma_z(x) &= \text{diag}(\sigma^2_z(x)) \\
    \Sigma_x(z) &= \sigma^2_x(z)I
\end{align*}

Restricting the $\Sigma_z$ to diagonal matrix induces orthogonality between latent channels~\cite{rolinek2019pca,kunin2019loss,lucas2019don}, which helps latent disentanglement and constrains the computation to be linear in $\dim{z}$. However, it is argued that this unduly limits the expressive power of encoder~\cite{shekhovtsov2021vaeapprox,wipf2023marginalization}.

The $\Sigma_x$ is usually assumed to be scalar and constant. The typical VAE that outputs only the mean $\mu_x$ is correspond to the case as it implies $\sigma^2_x=1/2$. This makes computation easier and avoids the optimization problem~\cite{rezende2018taming,mattei2018leveraging} that occurs when $\Sigma_x$ is a trainable parameter. The learnable $\Sigma_x$ tends to approach 0 as training progresses, causing the objective to diverge to infinite. However, the constant scalar variance does not allow VAE to reach the optimal latent structure, whereas the learnable scalar variance does~\cite{dai2018diagnosing,dai2021value}. This theoretical achievement is extended to the empirical nonlinear case~\cite{koehler2021variational,mattei2018leveraging}, which reports its superior performance despite being unstable and prone to overfitting. We will adopt scalar $\Sigma_x=\sigma^2_x(z)I$ but discuss constant $\sigma^2_x$.

\textbf{Learnable decoder variance.} The learnable variance of decoder $\sigma^2_x$ outperforms constant scalar variance~\cite{mattei2018leveraging,rybkin2021simple}, but introduces a nontrivial optimization problem~\cite{rezende2018taming,mattei2018leveraging}. In many conventional studies and implementations, the variance of decoder is often left constant. This empirically leads to degraded results~\cite{dai2021value,rybkin2021simple}, as the trainable variance has been discussed as essential for the optimization of Gaussian VAE~\cite{dai2018diagnosing,koehler2021variational,dai2021value}. Although few works successfully employed learnable variance stably~\cite{takahashi2018student,rybkin2021simple}, constant variance has been used in most prior research because learnable variance makes training process unstable and the effect is considered trivial~\cite{rybkin2021simple}.

\textbf{Beta-VAE.} Beta-VAE~\cite{higgins2016beta} on which applied works rather focus, demonstrates a simple yet effective enhancement on VAE. It introduces hyperparameter $\beta$ into the ELBO that balances the reconstruction error and KL divergence as shown in Equation~\eqref{elbo:beta}. $\beta$ influences the regularization by the KL divergence and latent disentanglement~\cite{higgins2016beta,burgess2018understanding,chen2018isolating,esmaeili2019structured}, which results in the efforts of fine-tuning $\beta$ in practice~\cite{kohl2018segmentation,pong2020skew}. These effects are attributed by estimating how well the variational distribution $q_\phi(z|x)$ follows the prior $p(z)$ in many cases~\cite{hoffman2017beta}.

\begin{align} 
\begin{split}
    L_\beta(\theta,\phi,x_i) = &-E_{z\sim q_\phi(z|x_i)}[\log{p_\theta(x|z)}] \\
    &+ \beta D_{KL}(q_\phi (z|x_i)||p(z))
\label{elbo:beta}
\end{split}
\end{align}

\textbf{Rate-distortion theory on $\beta$.} The balance of $\beta$ is explained by rate-distortion theory~\cite{huang2020evaluating} in which VAE is analogous to lossy compression~\cite{alemi2018fixing,bozkurt2021rate,bae2022multi}. The function of VAE is viewed as compressing a given $x$ into a usually lower-dimensional $z$ and restoring it, resembling a lossy compression system. In this context, reconstruction error corresponds to distortion and KL divergence term corresponds to rate in information theory. Therefore beta-VAEs are depicted by rate-distortion curve where each $\beta$ value determines a specific point. This indicates that beta-VAE changes the generation performance with $\beta$, unlike vanilla VAE, as the location of a point on the curve characterizes the model's performance.

\section{Beta-Sigma VAE}

\subsection{Categorizing Blurriness}

VAE is notorious for producing undesirable blurry output, which is a drawback given that its competitors, such as GAN~\cite{goodfellow2014gan} or diffusion model~\cite{sohl2015dpm}, produce very sharp output. Here, blurry means losing fine details that are usually present in high frequencies. This is a complex mix of phenomena, making it difficult to pinpoint a technical source. To ease further analysis, we categorize the blurry output problem into two types: poor reconstruction and poor generation.

Poor reconstruction refers to a model failing to reconstruct the training data regardless of generation. It corresponds to underfitting in general terms, which means that the VAE is not trained well, i.e., its likelihood for training or test data is low. The main cause is inadequate distribution modeling that does not fit the given data. In Gaussian VAE, the value of variance $\sigma_x$ and whether $\sigma_x$ is constant or learnable are important for good reconstruction. The impact of variance modeling has been reported extensively~\cite{dai2018diagnosing,rybkin2021simple,dai2021value}. For example, the low variance provides a high likelihood and thus improves reconstruction practically. The other cause is the limitation of neural network architecture, which is not the focus of this work, so many architectures and techniques have been proposed to address it~\cite{sonderby2016ladder,razavi2019vq2,child2020vdvae}.

Poor generation refers to a model failing to generate while being good at reconstruction relatively. In general terms, this corresponds to overfitting, but note that it is an evaluation of output generated from the prior $p(z)$, not the reconstruction of test data. It thus has little to do with likelihood. This is mainly due to the mismatch between the prior $p(z)$ and the aggregated posterior $q_\phi(z)=\int q_\phi(z|x)p(x)dx$, i.e., the gap between sampling in evaluation and reconstruction in training. To solve this, different choices of the distribution of the prior~\cite{tomczak2018vamp} or hierarchical VAE~\cite{dai2018diagnosing} have been introduced, but the simplest is beta-VAE~\cite{higgins2016beta}. Beta-VAE increases the influence of KL divergence as in Equation~\eqref{elbo:beta}, so that $q(z|x)$ matches $p(z)$ even if the parameter deviates from the optimal likelihood. This is a good way to resolve poor sampling because it helps to approach $q_\phi(z)=p(z)$ practically~\cite{burgess2018understanding}.

\begin{figure*}[t]
        \centering
        \includegraphics[width=\textwidth]{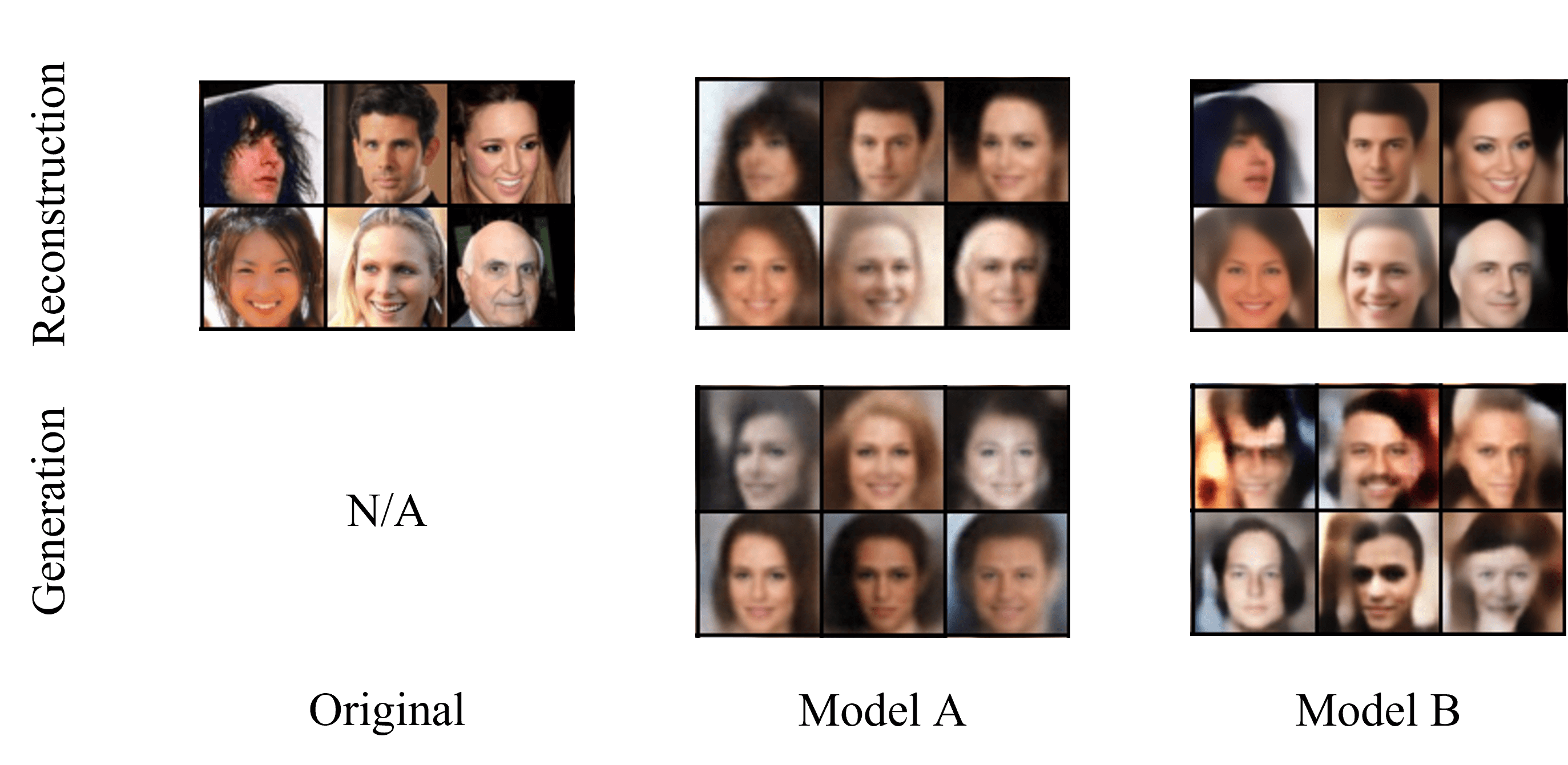}
\caption{The toy example of poor reconstruction and poor generation on CelebA dataset~\cite{liu2015celeba}. Model A displays a blurry reconstruction, but the quality of reconstruction and generation is consistent. Model B shows a relatively clear reconstruction, but the generation is blurry and unrealistic. Their setup is identical to the one in the experiment, and the samples are selected without any intention, i.e., no cherry picking.}
\label{fig:toy-example}
\end{figure*}

We provide the example in Fig.~\ref{fig:toy-example}. Model A is an example of poor reconstruction, trained with constant $\sigma^2_x$ and high $\beta$ $(=10)$. This model shows low likelihood, but the quality of reconstruction and generation is consistent. Model B is an example of poor generation, adopting learnable $\sigma^2_x$ without $\beta$ $(=1)$. This model demonstrates high likelihood, but the generation is relatively blurry and unrealistic. Their setup is identical to the one in the experiment. Since a model can only do one side well, we must distinguish between the two when approaching the blurry output problem.

\subsection{Problem Investigation}

Prior works and implementations practically assume constant variance building the decoder to output mean $\mu_x$~\cite{yu2020tutorial,rybkin2021simple}. This is problematic due to degraded performance and is further complicated by the introduction of $\beta$. We first explore the situation in which the variance and $\beta$ are equal. Specifying the distribution as Gaussian allows us to expand ELBO further. The reconstruction error, shown in Equation~\eqref{elbo:recon}, is expanded as Equation~\eqref{full-recon}.
\begin{align}
    -\log{p_\theta(x_i|z)} = \frac{(x_i-\mu_x(z))^2}{2\sigma^2_x(z)} + \frac{1}{2}\log{2\pi\sigma^2_x(z)}
\label{full-recon}
\end{align}
The log-sigma term on the right can be ignored in optimization if the variance is constant. Considering the beta-VAE with $\sigma^2_x=1/2$, then the $\beta$ of it mathematically equal to the $2\sigma^2_x$ in conventional VAE up to a constant multiplier~\cite{dai2018diagnosing,rybkin2021simple}, i.e., with a learning rate adaptation. This stems from the fact that the two objectives are identical in their form. Here we present a slightly more general relationship between $\beta$ and the variance in the same fashion, indicated in Equation~\eqref{lr-adapt} in which previously claimed equality is a special case of $C=1/2$.

\textbf{$\beta$ as constant decoder variance.} For the Gaussian beta-VAE with variance $\sigma^2_x=C$ and conventional Gaussian VAE with variance $\sigma^2_x=\beta \cdot C$ where $C$ is a constant scalar, the gradients of their objectives are identical up to a constant multiplier $\beta$, as indicated in Equation~\eqref{lr-adapt}. Hence, they are the same model in terms of neural network training, and the last $\equiv$ symbol in Equation~\eqref{lr-adapt} implies this. Note the subtlety that $C$ on the left is the variance of beta-VAE, and $\sigma^2_x$ on the right is of a general VAE.
{
\allowdisplaybreaks
\begin{align}
\begin{split}
      &L_\beta(\theta,\phi,x_i,\sigma^2_x) \\
    =\quad &E_{z\sim q_\phi(z|x_i)}\left[-\log{p_\theta(x|z)}\right] + \beta D_{KL}(q_\phi (z|x_i)||p(z)) \\
    =\quad &E_{z\sim q_\phi(z|x_i)}\left[\frac{(x_i-\mu_x(z))^2}{2\sigma^2_x(z)} + \frac{1}{2}\log{2\pi\sigma^2_x(z)}\right] + \beta D_{KL}(q_\phi (z|x_i)||p(z)) \\
    =\quad &E_{z\sim q_\phi(z|x_i)}\left[(x_i-\mu_x(z))^2\right]/2\sigma^2_x + \beta D_{KL}(q_\phi (z|x_i)||p(z)) + O(\log{\sigma^2_x})
    \nonumber
\end{split}
\\[3ex]
\begin{split}
    &-\text{ELBO}(\theta,\phi,x_i,\sigma^2_x) \\
    =\quad &E_{z\sim q_\phi(z|x_i)}\left[-\log{p_\theta(x|z)}\right] + D_{KL}(q_\phi (z|x_i)||p(z)) \\
    =\quad &E_{z\sim q_\phi(z|x_i)}\left[\frac{(x_i-\mu_x(z))^2}{2\sigma^2_x(z)} + \frac{1}{2}\log{2\pi\sigma^2_x(z)}\right] + D_{KL}(q_\phi (z|x_i)||p(z)) \\
    =\quad &E_{z\sim q_\phi(z|x_i)}\left[(x_i-\mu_x(z))^2\right]/2\sigma^2_x + D_{KL}(q_\phi (z|x_i)||p(z)) + O(\log{\sigma^2_x})
    \nonumber
\end{split}
\\[3ex]
\begin{split}
    \Rightarrow\quad &\nabla L_\beta(\theta,\phi, x_i, C) = -\beta\nabla\text{ELBO}(\theta, \phi, x_i, \beta \cdot C) \\
    \Rightarrow\quad &\beta \cdot C \equiv \sigma^2_x
    \label{lr-adapt}
\end{split}
\end{align}
}

The only value we can set in beta-VAE is the integrated parameter $\beta \cdot C$, not separate $\beta$ or $\sigma^2_x$, as they compensate each other. It means that introducing $\beta$ has almost no effect beyond tuning $\sigma^2_x$ as long as we use constant decoder variance, since it is completely absorbed in the variance. This not only negates the performance gain of $\beta$ but also makes the likelihood inconsistent, blocking meaningful model analysis.

\begin{figure*}[t]
        \centering
        \includegraphics[width=\textwidth]{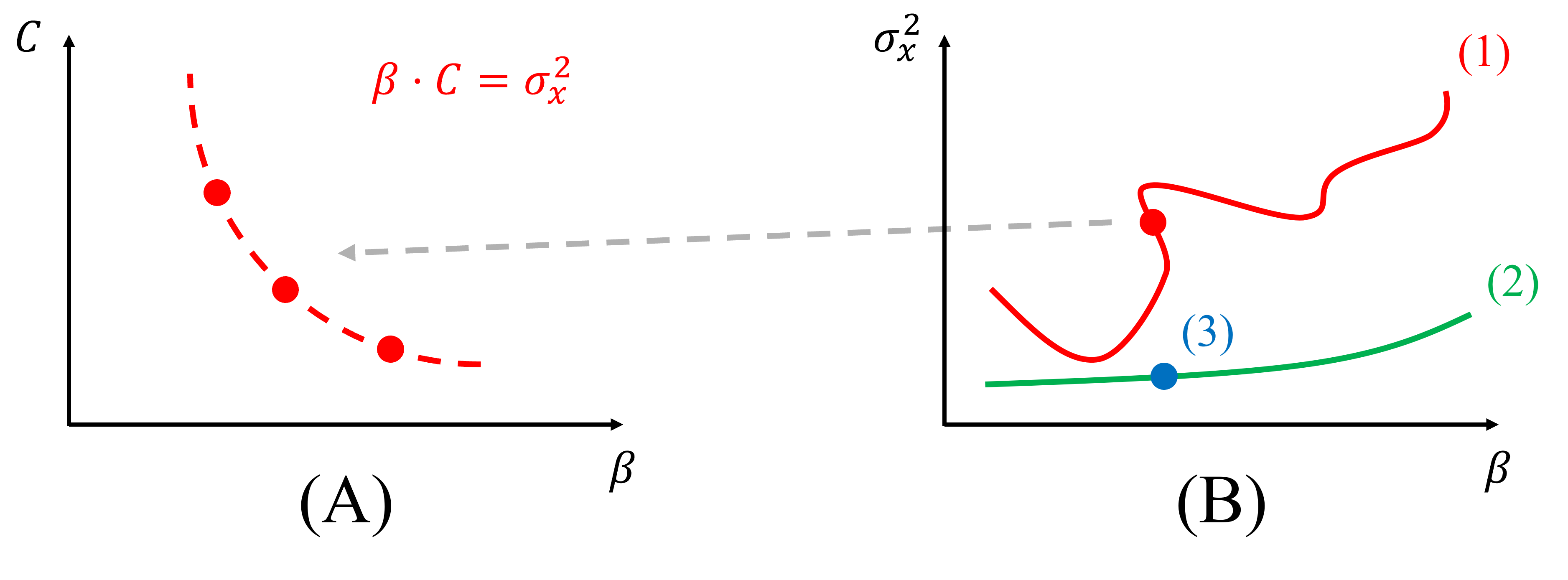}
\caption{The conceptual figure of optimizing $\sigma^2_x$ and $\beta$. \textbf{(A)} The dashed line indicates a constant $\sigma^2_x$ beta-VAE with same weights. Since the single integrated parameter $\beta \cdot C \equiv \sigma^2_x$ is set, researchers can arbitrarily choose $\beta$ and $C$ values for a $\sigma^2_x$. This harms VAE research by the inconsistency. \textbf{(B)} (1) A typical VAE cannot control each parameter. $\beta$ has almost no function beyond tuning $\sigma^2_x$ here. (2) Our method can tune the $\beta$ value while maintaining a reasonably low $\sigma^2_x$ value for the best likelihood. (3) The existing model with learnable decoder variance cannot adjust $\beta$, so it only represents a single point.}
\label{fig:beta-concept}
\end{figure*}

First, in the setting, decoder variance can be an arbitrary value. As given in Equation~\eqref{lr-adapt} and discussed in some works~\cite{dai2018diagnosing,rybkin2021simple}, if we consider the beta-VAE variance as $C=1/2$, then $\sigma^2_x=\beta/2$, leading to the consistent likelihood. However, most researchers treat $\beta$ as an isolated hyperparameter and calculate the likelihood from the beta-VAE variance $C$. This leaves the variance value to the researcher's discretion, as indicated in Fig.~\ref{fig:beta-concept}A. Consequently, studies that describe $\beta$ without specifying $C$ or code are not reproducible.

Worse still, the arbitrary variance introduces uncertainty in likelihood, since the reconstruction error is determined by $\sigma^2_x$ as in Equation~\eqref{full-recon}. This causes critical confusion in model analysis because the likelihood, which is a key value in the maximum likelihood estimation model, becomes inconsistent. For instance, constant variance beta-VAE has been usually considered as either $C=1/2$ or $C=\beta/2$ for the model with the same objective, or even parameters. The (lower bound of) log-likelihoods in each setting can be drastically different, so VAE studies that exhibit similar human-perceptual performance often show likelihood from $-10^6$ to $10^6$, making comparison virtually impossible.

Also, it is important to note that the goals of beta-VAE are different from those of conventional VAE. The beta-VAE is not the technique for obtaining the highest likelihood, but rather securing disentanglement or quality generation~\cite{higgins2016beta,burgess2018understanding,chen2018isolating,esmaeili2019structured}. It is evident from the very introduction of $\beta$, which makes the objective no longer the likelihood as in Equation~\eqref{elbo:beta}. However, the gradient of the constant $\sigma^2_x$ model is still within the likelihood, as demonstrated in Equation~\eqref{lr-adapt}. It does not lead to the benefits that only $\beta$ can achieve. Only the integrated parameter $\beta \cdot C \equiv \sigma^2_x$ is set, preventing control of each parameter. In this context, the role of $\beta$ is limited to adjusting $\sigma^2_x$, and the optimality of $\sigma^2_x$ and $\beta$ cannot be achieved. We depict it in Fig.~\ref{fig:beta-concept}A and Fig.~\ref{fig:beta-concept}B-1.

This inseparability of the variance and $\beta$ have confounded their respective effect. For example, researchers pursuing sharp generation ought to reduce the variance to increase likelihood~\cite{yu2020tutorial}. Many implementations, in fact, have chosen small $\beta$s (indeed, $\beta \cdot C \equiv \sigma^2_x$) to diminish the blurriness of generation. The optimal $\sigma^2_x$ and the optimal $\beta$ are different. The optimal $\sigma^2_x$ is arguably the maximizer of likelihood, but the optimal $\beta$ depends on the purpose. In~\cite{rybkin2021simple} dealing with similar confusion, they have pointed out the pervasive imprecise implementation of $\sigma^2_x$, but their claim that the optimal $\sigma^2_x$ is also the optimal $\beta$ is incorrect. Such confusion not only harms the practical performance of VAE but also the theoretical analysis of VAE.

A natural approach to address the limitation of integrated parameter $\beta \cdot C \equiv \sigma^2_x$ is to separate the two parameters. Since the constant variance beta-VAE cannot achieve the aim, we employ the learnable variance beta-VAE. Still, implementing the learnable decoder variance poses an optimization problem~\cite{rezende2018taming}. We first analyze how the objective behaves in the setting.

When the variance of decoder is considered as the trainable parameter, $\sigma^2_x$ and $\beta$ are distinct to each other, as the gradient of the objective changes. The key to the distinction is the log-sigma term in Equation~\eqref{full-recon}. In this setting, Equation~\eqref{lr-adapt} does not hold since the log-sigma term is not constant. The log-sigma term is derived from the normalizing factor of Gaussian probability density function, allowing the decoder function to remain as a probability distribution. Letting the variance change rather than constant enhances the expressiveness of model, but the distribution becomes uncontrollable if the variance converges to $0$ or $\infty$.

In optimization, the log-sigma term prevents the infinitely large $\sigma^2_x$ to reduce the objective~\cite{lucas2019don}. A large variance compensates for the error arising from prediction failure, as illustrated in Equation~\eqref{full-recon}, hence $\sigma^2_x$ may diverge to infinity without the log-sigma term. Namely, the log-sigma term encourages the model to learn a large $\sigma^2_x$ for challenging samples and a small $\sigma^2_x$ for easier ones. Consequently, the variance represents an uncertainty, making it reasonable that its value decreases as training progresses, even if it approaches 0. This leads to the unstable optimization caused by the zero variance. Indeed, it has been claimed that this infinite gradient helps in achieving the optimal latent structure~\cite{dai2021value}.

Although it intuitively or theoretically makes sense, unstable optimization is undesirable for practical uses. A few works~\cite{takahashi2018student,rybkin2021simple} have provided implementations for the stable decoder with learnable variance exploiting the property of Gaussian, which we employ in our method.

\subsection{Method}

We propose a method to separate the variance of decoder and $\beta$, simply introducing $\beta$ with learnable variance. To maintain stable optimization, we first adopt the optimal variance.

\textbf{Optimal decoder variance $\sigma^2_x$.} For the reconstruction error of a Gaussian VAE (Equation~\eqref{full-recon}), a single sample $x_i$, and its sampled latent $z_i$, we can find a analytical optimal $\sigma^{2^*}_x(z_i)$ for a given $(x_i-\mu(z_i))^2$. 
\begin{align*}
    &\frac{\partial}{\partial \sigma_x} [-\text{ELBO}(\theta, \phi, x_i, z_i) ]\\
    =\quad&\frac{\partial}{\partial \sigma_x}[ -\log{p_\theta(x_i|z_i)} + D_{KL}(q_\phi (z|x_i)||p(z)) ] \\
    =\quad&\frac{\partial}{\partial \sigma_x}[\frac{(x_i-\mu_x(z_i))^2}{2\sigma^2_x(z_i)} + \frac{1}{2}\log{2\pi\sigma^2_x(z_i)} + O(1)] \\
    =\quad& -\frac{(x_i-\mu_x(z_i))^2}{\sigma^3_x(z_i)} + \frac{1}{\sigma_x(z_i)}\quad =\quad 0 \\
    \Rightarrow\quad& \sigma^{2^*}_x(z_i) = (x_i-\mu_x(z_i))^2
\end{align*}
This is an alternative to directly implementing trainable variance~\cite{yu2020tutorial,rybkin2021simple}. We employ this because it is mathematically clear and easy to implement.

Albeit it has been argued as the method to find the optimal $\beta$~\cite{rybkin2021simple}, according to our claim, the optimal $\sigma^2_x$ is not identical to the optimal $\beta$. Rather, the Gaussian VAE with optimal decoder variance is not associated with $\beta$, i.e., $\beta=1$, as demonstrated in Fig.~\ref{fig:beta-concept}B-3. $\sigma^2_x$ and $\beta$ should be taken as different parameters.

\begin{align}
\begin{split}
    L_{\beta\sigma}(\theta,\phi) =\ &\frac{1}{2}E_{z\sim q_\phi(z|x)}[\log{2\pi(x-\mu_x(x))^2} + 1] \\
    &+ \beta D_{KL}(q_\phi (z|x)||p(z))
\label{loss_ours}
\end{split}
\end{align}

Then $\beta$ can be reintroduced into the optimal $\sigma^2_x$ model. As a result, we build a new objective named Beta-Sigma VAE (BS-VAE) as shown in Equation~\eqref{loss_ours}. Although it looks like a straightforward and simple extension, BS-VAE achieves the control of each parameter, as illustrated in Fig.~\ref{fig:beta-concept}B-2. It takes advantage of both parameters and ensures that the same model and weights always provide the same likelihood value. It also shows significant performance improvement over prior works in our experimental evaluation.

\begin{figure*}[t]
        \centering
        \includegraphics[width=\textwidth]{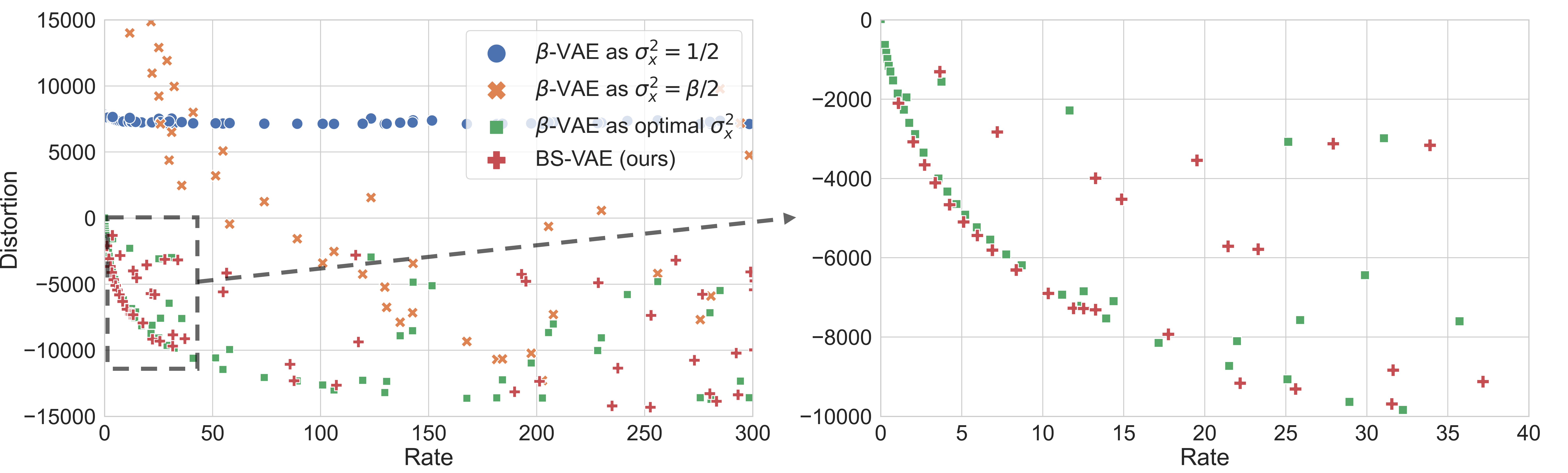}
\caption{The rate-distortion curve plotting BS-VAEs and conventional beta-VAEs with constant $\sigma^2_x$. The constant variance can be interpreted in various ways, so the optimal $\sigma^2_x$ that leads distortion to the lower bound and two common $\sigma^2_x$s are indicated. BS-VAEs outperform the conventional models by any interpretation of $\sigma^2_x$.}
\label{fig:rd-curve}
\end{figure*}

\section{Experimental Evaluation}

\subsection{Evaluation Setup}

We train and compare BS-VAEs and typical beta-VAEs with constant $\sigma^2_x$, which provide empirical evidence of our proposition. First, to reveal the ambiguity of reconstruction error, we visualize the rate-distortion curve, which exhibits the performance of each model as a point on the curve. The proposed BS-VAE draws a single curve. On the other hand, conventional beta-VAE with constant decoder variance has multiple interpretations along the fixed variance values and corresponding distortion of the curve. We test in three different ways: $\sigma^2_x=1/2$, $\sigma^2_x=\beta/2$, which are the views often adopted in previous research, and the case with optimal $\sigma^2_x$, which is the upper bound of beta-VAE performance interpretation. Secondly, we evaluate the VAEs based on proxy metrics, i.e. Fréchet inception distance~\cite{heusel2017fid} (FID) and log-likelihood on unseen data. Although likelihood is a good indicator of generative model and it directly measures the optimization of VAE, generation is difficult to be evaluated in a single figure. For example, a fully memorized model, i.e., a lossless compression system, achieves an infinite log-likelihood on training set, ignoring important values such as diversity. Thus the proxy metrics are convincing indicators by preventing the model from simply remembering the training data. To improve FID, $\beta$ of beta-VAE has been adjusted by practitioners at the cost of likelihood frequently. log-likelihood on unseen data has been used as an indicator for generalization capability in previous works~\cite{sonderby2016ladder,tomczak2018vamp}. Additionally, to evaluate generative neural networks, we conduct a qualitative evaluation of generated samples.

All models consist of a Gaussian encoder with diagonal covariance matrix and a Gaussian decoder. We employ common shallow convolutional neural network architecture with a residual connection to implement VAEs for our experiments. They are evaluated on popular computer vision datasets, CelebA~\cite{liu2015celeba} and MNIST~\cite{lecun1998mnist}. They consist of 4-layer residual block encoder and 4-layer convolutional decoder with 64 latent channels to train CelebA dataset. MNIST test networks are simplified to have 3 layers for each encoder and decoder with 32 latent channels. We train each model for 50 epochs using AdamW optimizer~\cite{loshchilov2018adamw} and evaluate them on the fully trained model. For more specific settings, see \url{https://github.com/overnap/BS-VAE}.

As the evaluation is for proof-of-concept, it is conducted on relatively shallow neural networks and light datasets. We emphasize that BS-VAE is applicable to most VAE-variants, because our argument is about the parameterization of Gaussian VAE, independent of architecture and scale. However, it is difficult to ensure that it applies to the larger architecture using VAE as a part, such as latent diffusion model~\cite{rombach2022ldm}. The discussion about it is an interesting future work.

\subsection{Experimental Results}

\begin{table}[b!]
{
\caption{Proxy metric evaluations of BS-VAEs and constant decoder variance beta-VAEs with various $\beta$s. The FID~\cite{heusel2017fid} and the log-likelihood on test set are shown with the common log-likelihood for reference. The models are trained five times each, showing their means. BS-VAE obtains the best likelihood at $\beta=1$ and the best FID at $\beta=10$, demonstrating that optimal $\sigma^2_x$ does not mean optimal $\beta$. }
\centering
\sisetup{separate-uncertainty}
\begin{tabular}{|c|c||c|c|c|c|c|c|}
    \hline
    \multicolumn{2}{|c||}{Model} & \multicolumn{3}{c|}{CelebA} & \multicolumn{3}{c|}{MNIST} \\
    \hline
    {Name} & {$\beta$} & {FID ($\downarrow$)} & {Test $\log{p_\theta(x)}$} & {$\log{p_\theta(x)}$} & {FID ($\downarrow$)} & {Test $\log{p_\theta(x)}$} & {$\log{p_\theta(x)}$} \\
    \hline
    \multirow{5}{5.5em}{Beta-VAE with\\ constant $\sigma^2_x$}
    & 0.01  & 194.7 & \textbf{> 10684} & \textbf{> 10762} & 190.3 & \textbf{> 667} & \textbf{> 659}  \\
    & 0.1   & 151.7 & > 10384 & > 10412 & \textbf{163.6} & > 626 & > 618  \\
    & 1     & \textbf{126.4} & > 10616 & > 10626 & 225.8 & > 291 & > 286  \\
    & 10    & 149.4 & > 6923   & > 6898   & 351.7  & > -19  & > -20   \\
    & 100   & 235.8 & > 2233     & > 2190    & 352.5  & > -19  & > -20   \\
    \hline
    \multirow{5}{5.5em}{BS-VAE (Ours)}
    & 0.01  & 188.5 & > 10772 & > 10848 & 67.4   & > 796 & > 788  \\
    & 0.1   & 130.2 & > 12996 & > 13037 & 75.5   & > 850 & > 840  \\
    & 1     & 90.8  & \textbf{> 14384} & \textbf{> 14434} & 59.2   & \textbf{> 887} & \textbf{> 877}  \\
    & 10    & \textbf{73.7}   & > 13205  & > 13256  & \textbf{38.4}   & > 662 & > 656  \\
    & 100   & 106.2 & > 7668  & > 7630  & 332.8  & > -15  & > -17   \\
    \hline
\end{tabular}\par
}
\label{table:metric}
\end{table}

We train VAEs on CelebA with $\beta$ scaled from 0.0001 to 1000, which is wide enough for common use. We evaluate ELBO of the models and plot their rate-distortion curves as summarized in Fig.~\ref{fig:rd-curve}. BS-VAEs (red crosses) outperform two types of constant variance beta-VAEs (blue circles and orange x). beta-VAEs as $\sigma^2_x=1/2$ appear to fall short in drawing the desired rate-distortion trade-off. In the $\sigma^2_x=\beta/2$ case, distortions are significantly high compared to our model in low rate cases. As the rate decreases, the performance gap between $\sigma^2_x=\beta/2$ case and ours becomes larger. This is a critical drawback of beta-VAEs since VAE naturally pursues to reduce the rate (i.e., KL divergence) in training to satisfy given tasks. This can be explained by Equation~\eqref{lr-adapt}. Assuming $C=1/2$, $\beta=2\sigma^2_x$ holds through learning rate adaptation. Extended to the trainable $\sigma^2_x$ VAE, the equation no longer holds, and the more delicate relationship between $\beta$ and $\sigma^2_x$ is disclosed by the same development.
\begin{equation*}
    \beta = 2\sigma^2_x(z) + \frac{\sigma^2_x(z)\log{2\pi\sigma^2_x(z)}}{D_{KL}(q_\phi (z|x)||p(z))}
\end{equation*}
Notably, the influence of the log-sigma term, governed by the KL divergence term in its denominator, increases as the KL divergence diminishes, explaining the performance gap clearly.

Proposed BS-VAEs outperform compared to the constant variance beta-VAEs as optimal $\sigma^2_x$ (green squares) in Fig.~\ref{fig:rd-curve}. The constant variance models evaluated with optimal $\sigma^2_x$ represent the upper bound for their likelihood. Therefore, BS-VAEs generally achieves better performance than typical beta-VAEs regardless of the interpretation of $\sigma^2_x$, by leveraging both parameters. Previous studies have shown similar results only at certain $\beta$, especially near the optimal $\sigma^2_x$ value~\cite{rybkin2021simple,dai2021value}.

We train VAEs with $\beta$ from 0.01 to 100 on CelebA and MNIST and present their proxy metrics in Table~\ref{table:metric}. The models are trained five times each, and the results are shown with their means. Note that ELBO is calculated instead of the direct log-likelihood. For constant $\sigma^2_x$ models, the lower bound of ELBO is shown for meaningful comparison, i.e., assuming optimal $\sigma^2_x$. Otherwise, there is much of a gap like the left of Fig.~\ref{fig:rd-curve}, e.g., $\log{p_\theta(x)} = -8000$.

In both datasets, BS-VAEs demonstrate better performance than constant $\sigma^2_x$ models where $\beta=1$. Note that lower FID and higher likelihood indicate better performance in the tasks. Furthermore, BS-VAEs with $\beta=1$ show better performance compared to the constant variance models over the entire $\beta$ range. These results concur with those reported in previous studies: BS-VAE with $\beta=1$ is conceptually identical to~\cite{dai2021value} and implementationally identical to~\cite{rybkin2021simple}. We thus claim that the improvement comes from the benefit of learnable decoder variance rather than any implementation-specific gain.

\begin{figure*}[t]
        \centering
        \includegraphics[width=\textwidth]{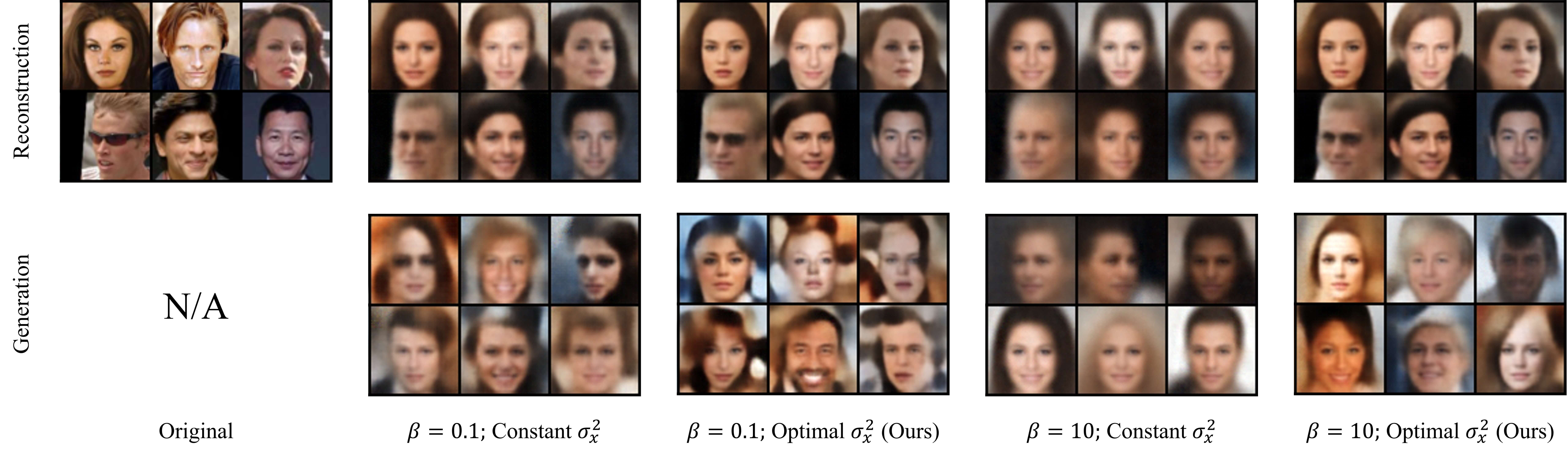}
\caption{Reconstructed or generated samples of common beta-VAEs with constant decoder variance and our BS-VAEs. Our models maintain good reconstruction quality within tested $\beta$s. The samples are selected without any intention, i.e., no cherry picking.}
\label{fig:sample}
\end{figure*}

As illustrated in BS-VAEs with $\beta\neq 1$ in Table~\ref{table:metric}, we obtain learnable variance models with various $\beta$s by the reintroduction of $\beta$ into the optimal variance model. They all attain better FID scores compared to the constant models for the same $\beta$. As the good proxy metric is a goal of tuning $\beta$, the empirical best $\beta$ for our model is 10, exhibiting significant performance gain. This naturally disproves the previous claim that the optimal $\sigma^2_x$ means the optimal $\beta$~\cite{rybkin2021simple}. Even in the optimal variance model, $\beta$ can be adjusted to achieve better proxy metrics or latent disentanglement. Moreover, BS-VAEs attain the best likelihood at $\beta=1$ where the objective remains as likelihood. This is not the case in constant models where the likelihood increases as $\beta$ decreases despite the objective drifting away from the log-likelihood. These results align with our arguments in Section 3 and Fig.~\ref{fig:beta-concept}.

We display reconstructed and generated samples of these models in Fig.~\ref{fig:sample}. Arguably, BS-VAEs excel in reconstruction quality regardless of the $\beta$ value, meeting the basic purpose of VAE, i.e., lossy compression. A possible explanation for this is that moderate $\beta$ values do not hinder the achievement of optimal latent structure~\cite{dai2021value}. In BS-VAE, varying $\beta$ only changes generation quality, while the conventional VAE does not. This is because the $\beta$ we adjust in the constant model, as shown in Equation~\eqref{lr-adapt} and Fig.~\ref{fig:beta-concept}A, is actually the integrated parameter $\beta \cdot C \equiv \sigma^2_x$. BS-VAE at $\beta=10$ exploits both $\sigma^2_x$ and $\beta$, resulting in both good reconstruction and good generation.

\section{Conclusion}
We investigated and addressed the blurry output problem of VAE. In particular, we elucidated the confusion between the variance of Gaussian decoder $\sigma^2_x$ and $\beta$ of beta-VAE~\cite{higgins2016beta}. We also proposed BS-VAE to handle the indistinguishability problem of beta-VAE with constant decoder variance. Our BS-VAE is simple but explicitly separates the $\sigma^2_x$ and $\beta$, demonstrating competitive performance over prior work with predictable and meaningful analysis. We expect that the following research avoids ambiguity and obtains optimal VAE performance in applications.

\subsubsection{Acknowledgements} This work was supported by the National Research Foundation of Korea (NRF) grant (No.NRF-2020R1A2C1015146) and the IITP (Institute of Information \& Communications Technology Planning \& Evaluation) grant (National Program for Excellence in SW, 2023-0-00042 in 2024) funded by the Korea government (Ministry of Science and ICT).

%
%
%
\bibliographystyle{splncs04}
\bibliography{mybibliography}

\begin{thebibliography}{10}
\providecommand{\url}[1]{\texttt{#1}}
\providecommand{\urlprefix}{URL }
\providecommand{\doi}[1]{https://doi.org/#1}

\bibitem{alemi2018fixing}
Alemi, A., Poole, B., Fischer, I., Dillon, J., Saurous, R.A., Murphy, K.: Fixing a broken elbo. In: ICML. pp. 159--168. PMLR (2018)

\bibitem{bae2022multi}
Bae, J., Zhang, M.R., Ruan, M., Wang, E., Hasegawa, S., Ba, J., Grosse, R.B.: Multi-rate vae: Train once, get the full rate-distortion curve. In: ICLR (2023)

\bibitem{bowman2016text}
Bowman, S.R., Vilnis, L., Vinyals, O., Dai, A., Jozefowicz, R., Bengio, S.: Generating sentences from a continuous space. In: Proceedings of The 20th SIGNLL CoNLL. p.~10. Association for Computational Linguistics (2016)

\bibitem{bozkurt2021rate}
Bozkurt, A., Esmaeili, B., Tristan, J.B., Brooks, D., Dy, J., van~de Meent, J.W.: Rate-regularization and generalization in variational autoencoders. In: AISTATS. pp. 3880--3888. PMLR (2021)

\bibitem{burgess2018understanding}
Burgess, C.P., Higgins, I., Pal, A., Matthey, L., Watters, N., Desjardins, G., Lerchner, A.: Understanding disentangling in $\beta$-vae. arXiv preprint arXiv:1804.03599  (2018)

\bibitem{chen2018isolating}
Chen, R.T., Li, X., Grosse, R.B., Duvenaud, D.K.: Isolating sources of disentanglement in variational autoencoders. NeurIPS  \textbf{31} (2018)

\bibitem{child2020vdvae}
Child, R.: Very deep vaes generalize autoregressive models and can outperform them on images. arXiv preprint arXiv:2011.10650  (2020)

\bibitem{dai2021value}
Dai, B., Wenliang, L., Wipf, D.: On the value of infinite gradients in variational autoencoder models. NeurIPS  \textbf{34},  7180--7192 (2021)

\bibitem{dai2018diagnosing}
Dai, B., Wipf, D.: Diagnosing and enhancing vae models. In: ICLR (2018)

\bibitem{duan2023compress2}
Duan, Z., Lu, M., Ma, Z., Zhu, F.: Lossy image compression with quantized hierarchical vaes. In: Proceedings of the IEEE/CVF WACV. pp. 198--207 (2023)

\bibitem{esmaeili2019structured}
Esmaeili, B., Wu, H., Jain, S., Bozkurt, A., Siddharth, N., Paige, B., Brooks, D.H., Dy, J., Meent, J.W.: Structured disentangled representations. In: AISTATS. pp. 2525--2534. PMLR (2019)

\bibitem{goodfellow2014gan}
Goodfellow, I., Pouget-Abadie, J., Mirza, M., Xu, B., Warde-Farley, D., Ozair, S., Courville, A., Bengio, Y.: Generative adversarial nets. NeurIPS  \textbf{27} (2014)

\bibitem{heusel2017fid}
Heusel, M., Ramsauer, H., Unterthiner, T., Nessler, B., Hochreiter, S.: Gans trained by a two time-scale update rule converge to a local nash equilibrium. NeurIPS  \textbf{30} (2017)

\bibitem{higgins2016beta}
Higgins, I., Matthey, L., Pal, A., Burgess, C., Glorot, X., Botvinick, M., Mohamed, S., Lerchner, A.: beta-vae: Learning basic visual concepts with a constrained variational framework. In: ICLR (2016)

\bibitem{hoffman2017beta}
Hoffman, M.D., Riquelme, C., Johnson, M.J.: The $\beta$-vae’s implicit prior. In: Workshop on Bayesian Deep Learning, NIPS. pp.~1--5 (2017)

\bibitem{huang2020evaluating}
Huang, S., Makhzani, A., Cao, Y., Grosse, R.: Evaluating lossy compression rates of deep generative models. In: ICML. pp. 4444--4454. PMLR (2020)

\bibitem{kingma2013auto}
Kingma, D.P., Welling, M.: Auto-encoding variational bayes. arXiv preprint arXiv:1312.6114  (2013)

\bibitem{koehler2021variational}
Koehler, F., Mehta, V., Zhou, C., Risteski, A.: Variational autoencoders in the presence of low-dimensional data: landscape and implicit bias. arXiv preprint arXiv:2112.06868  (2021)

\bibitem{kohl2018segmentation}
Kohl, S., Romera-Paredes, B., Meyer, C., De~Fauw, J., Ledsam, J.R., Maier-Hein, K., Eslami, S., Jimenez~Rezende, D., Ronneberger, O.: A probabilistic u-net for segmentation of ambiguous images. NeurIPS  \textbf{31} (2018)

\bibitem{kunin2019loss}
Kunin, D., Bloom, J., Goeva, A., Seed, C.: Loss landscapes of regularized linear autoencoders. In: ICML. pp. 3560--3569. PMLR (2019)

\bibitem{lecun1998mnist}
LeCun, Y., Bottou, L., Bengio, Y., Haffner, P.: Gradient-based learning applied to document recognition. Proceedings of the IEEE  \textbf{86}(11),  2278--2324 (1998)

\bibitem{liu2015celeba}
Liu, Z., Luo, P., Wang, X., Tang, X.: Deep learning face attributes in the wild. In: Proceedings of the IEEE ICCV. pp. 3730--3738 (2015)

\bibitem{loshchilov2018adamw}
Loshchilov, I., Hutter, F.: Decoupled weight decay regularization. In: ICLR (2018)

\bibitem{lucas2019don}
Lucas, J., Tucker, G., Grosse, R.B., Norouzi, M.: Don't blame the elbo! a linear vae perspective on posterior collapse. NeurIPS  \textbf{32} (2019)

\bibitem{mattei2018leveraging}
Mattei, P.A., Frellsen, J.: Leveraging the exact likelihood of deep latent variable models. NeurIPS  \textbf{31} (2018)

\bibitem{nair2018reinforcement}
Nair, A.V., Pong, V., Dalal, M., Bahl, S., Lin, S., Levine, S.: Visual reinforcement learning with imagined goals. NeurIPS  \textbf{31} (2018)

\bibitem{norouzi2020augmentation}
Norouzi, S., Fleet, D.J., Norouzi, M.: Exemplar vae: Linking generative models, nearest neighbor retrieval, and data augmentation. NeurIPS  \textbf{33},  8753--8764 (2020)

\bibitem{pong2020skew}
Pong, V.H., Dalal, M., Lin, S., Nair, A., Bahl, S., Levine, S.: Skew-fit: state-covering self-supervised reinforcement learning. In: ICML. pp. 7783--7792 (2020)

\bibitem{razavi2019vq2}
Razavi, A., Van~den Oord, A., Vinyals, O.: Generating diverse high-fidelity images with vq-vae-2. NeurIPS  \textbf{32} (2019)

\bibitem{rezende2014stochastic}
Rezende, D.J., Mohamed, S., Wierstra, D.: Stochastic backpropagation and approximate inference in deep generative models. In: ICML. pp. 1278--1286. PMLR (2014)

\bibitem{rezende2018taming}
Rezende, D.J., Viola, F.: Taming vaes. arXiv preprint arXiv:1810.00597  (2018)

\bibitem{rolinek2019pca}
Rolinek, M., Zietlow, D., Martius, G.: Variational autoencoders pursue pca directions (by accident). In: Proceedings of the IEEE/CVF CVPR. pp. 12406--12415 (2019)

\bibitem{rombach2022ldm}
Rombach, R., Blattmann, A., Lorenz, D., Esser, P., Ommer, B.: High-resolution image synthesis with latent diffusion models. In: Proceedings of the IEEE/CVF CVPR. pp. 10684--10695 (2022)

\bibitem{rybkin2021simple}
Rybkin, O., Daniilidis, K., Levine, S.: Simple and effective vae training with calibrated decoders. In: ICML. pp. 9179--9189. PMLR (2021)

\bibitem{shekhovtsov2021vaeapprox}
Shekhovtsov, A., Schlesinger, D., Flach, B.: Vae approximation error: Elbo and exponential families. In: ICLR (2021)

\bibitem{sohl2015dpm}
Sohl-Dickstein, J., Weiss, E., Maheswaranathan, N., Ganguli, S.: Deep unsupervised learning using nonequilibrium thermodynamics. In: ICML. pp. 2256--2265. PMLR (2015)

\bibitem{sonderby2016ladder}
S{\o}nderby, C.K., Raiko, T., Maal{\o}e, L., S{\o}nderby, S.K., Winther, O.: Ladder variational autoencoders. NeurIPS  \textbf{29} (2016)

\bibitem{takahashi2018student}
Takahashi, H., Iwata, T., Yamanaka, Y., Yamada, M., Yagi, S.: Student-t variational autoencoder for robust density estimation. In: IJCAI. pp. 2696--2702 (2018)

\bibitem{tomczak2018vamp}
Tomczak, J., Welling, M.: Vae with a vampprior. In: AISTATS. pp. 1214--1223. PMLR (2018)

\bibitem{wipf2023marginalization}
Wipf, D.: Marginalization is not marginal: no bad vae local minima when learning optimal sparse representations. In: ICML. pp. 37108--37132. PMLR (2023)

\bibitem{yu2020tutorial}
Yu, R.: A tutorial on vaes: From bayes' rule to lossless compression. arXiv preprint arXiv:2006.10273  (2020)

\end{thebibliography}
%




\end{document}